\pdfoutput=1

\documentclass[11pt]{article}

\usepackage{emnlp2021}
\usepackage{graphicx}
\usepackage{times}
\usepackage{latexsym}
\usepackage{bbm}
\usepackage{subfigure}
\usepackage{amsmath}
\usepackage{url}

\usepackage[T1]{fontenc}

\usepackage[utf8]{inputenc}

\usepackage{microtype}
\usepackage{booktabs}

\title{Revisiting Self-Training for Few-Shot Learning of Language Model}

\author{Yiming Chen$^{\dag,\ddag}$ \quad Yan Zhang$^{\dag}$ \quad Chen Zhang$^{\dag}$ \quad Grandee Lee$^{\dag}$ \\ \quad \textbf{Ran Cheng$^{{\ddag},}$\thanks{\quad Corresponding author.}} \quad \textbf{Haizhou Li$^{\dag, \star, \star\star}$} \\
        $^\dag$National University of Singapore~~$^\ddag$Southern University of Science and Technology \\
        $^{\star}$The Chinese University of Hong Kong (Shenzhen) \quad $^{\star\star}$Kriston AI Lab, China\\
        \tt \{yiming.chen,chen\_zhang,grandee.lee\}@u.nus.edu, \\
        \tt \{haizhou.li,eleyanz\}@nus.edu.sg, \\
        \tt ranchengcn@gmail.com

}

\begin{document}
\maketitle

\begin{abstract}

As unlabeled data carry rich task-relevant information, they are proven useful for few-shot learning of language model. The question is how to effectively make use of such data. In this work, we revisit the self-training technique for language model fine-tuning and present a state-of-the-art prompt-based few-shot learner, SFLM. Given two views of a text sample via weak and strong augmentation techniques, SFLM generates a pseudo label on the weakly augmented version. Then, the model predicts the same pseudo label when fine-tuned with the strongly augmented version. This simple approach is shown to outperform other state-of-the-art supervised and semi-supervised counterparts on six sentence classification and six sentence-pair classification benchmarking tasks. In addition, SFLM only relies on a few in-domain unlabeled data. We conduct a comprehensive analysis to demonstrate the robustness of our proposed approach under various settings, including augmentation techniques, model scale, and few-shot knowledge transfer across tasks. \footnote{Our code is publicly available at \url{https://github.com/MatthewCYM/SFLM}}

\end{abstract}
\section{Introduction}
Pre-trained language models~\citep{devlin-etal-2019-bert,Liu2019RoBERTaAR,radford2019language,yang2019xlnet,Lan2020ALBERTAL,Raffel2020ExploringTL,clark2020electra} have set new state-of-the-art performance in many downstream NLP tasks. However, such performance often relies on large-scale high-quality supervision. Unfortunately, labeled data are not always available in practice. 

Recently, ~\citet{Brown2020LanguageMA} study how to facilitate the few-shot learning of language models via the GPT-3 model. It achieves remarkable performance on many NLP datasets without any gradient updates, by incorporating task-specific prompts into the text and reformulating the task as language modeling problems. However, GPT-3 has 175B parameters, that has a footprint too large for many real-world applications.  ~\citet{gao2020making} applies the concept of prompt strategies in GPT-3 to small-footprint language models, such as BERT~\citep{devlin-etal-2019-bert} and RoBERTa~\citep{Liu2019RoBERTaAR}. After fine-tuned on a few annotated samples, the small-footprint models exhibit comparable performance to that of  the large-footprint GPT-3 model.  However, the performance of these models is still lagging behind those under supervised learning, which have a much smaller footprint. Intuitively, unlabeled data also carry rich information of downstream tasks and are more available than labelled data. In this paper, we focus on the few-shot learning of language model with a small amount of labeled and unlabeled data.

Semi-supervised learning benefits from partially labeled datasets. A common implementation of semi-supervised learning is self-training,  which leverages supervision signals offered by labeled data to create pseudo-labels for unlabeled data. These pseudo labels serve as additional supervision to refine the models~\citep{yarowsky-1995-unsupervised,Blum1998CombiningLA,Zhu2005SemiSupervisedLL,qiu-etal-2019-graph, Zoph2020RethinkingPA}. Recent works~\citep{schick2020s,schick-schutze-2021-exploiting} apply self-training to language model few-shot learning in an iterative manner whereby multiple generations of models are trained on data pseudo-labeled by the ensemble of previous generations.  However, the amount of in-domain unlabeled data required by these methods is quite large, that limits the scope of the applications,   especially for low-resource downstream tasks.~\citet{Du2020SelftrainingIP} try to retrieve more task-relevant unlabeled data from open-domain corpus, but the method depends on a quality sentence encoder. 

To better address the above issue,  we revisit the \textbf{S}elf-training techniques and introduce a data-efficient \textbf{F}ew-shot learner of \textbf{L}anguage \textbf{M}odel (SFLM). Inspired by recent advances in semi-supervised representation learning for images~\citep{Sohn2020FixMatchSS}, SFLM combines pseudo-labeling with consistency regularization.

Next, we briefly describe the workflow. Given each unlabeled sentence, we construct two views through weak augmentation (random dropout) and strong augmentation (token masking), respectively. The weakly-augmented view is first passed to a prompt-based language model~\citep{gao2020making} to derive the pseudo-label, while the strongly-augmented view is passed through the model to predict the probability distribution over classes, which is compared to the pseudo-label to derive a cross-entropy loss.  This learning procedure encourages the model to capture the information that is almost outside of the data distribution, leading to effective utilization of data.
 
We evaluate SFLM on two groups of tasks -- sentence classification and sentence-pair classification. Experiments show that our model outperforms other supervised and semi-supervised baselines. We also conduct a detailed analysis of the data efficiency of our model by examining its performance w.r.t various ratios of the amount of the unlabeled data to that of the labelled data. We find out that the performance gain diminishes as more unlabeled data are used.  We further extend our method for a more challenging scenario: few-shot transfer across tasks, where the model is first trained on the labeled data of a source task and the unlabeled data of a target task, then evaluated on the target task. We provide the analysis of the factors that affect the model performance to motivate future research.

\section{Related Work}
\subsection{Few-Shot Learning of Language Model}
It is desirable to reduce the amount of labeled data for language model fine-tuning, a.k.a., language model few-shot learning. The popular methods usually address this problem with meta-learning~\citep{vinyals2016matching,snell2017prototypical,finn2017model}, which first pre-trains a model on a set of auxiliary tasks, then fine-tunes on the task of interest~\citep{yu-etal-2018-diverse,han-etal-2018-fewrel,Bao2020FewshotTC,bansal-etal-2020-self}. 

Recently,~\citet{Brown2020LanguageMA} proposes GPT-3 and demonstrates that the language model itself has a great potential for few-shot learning through task demonstrations and prompts. As GPT-3~\citep{Brown2020LanguageMA} has an extremely large footprint, that limits its scope of applications.

More recent studies explore few-shot learning with pre-trained language models~\citep{Gunel2020SupervisedCL,schick2020s,gao2020making} of smaller size. A representative example is the LM-BFF by~\citep{gao2020making}, which explores automatic prompt generation and prompt-based fine-tuning with a RoBERTa-large~\citep{Liu2019RoBERTaAR} language model in a few-shot setup. LM-BFF has achieved comparable results w.r.t methods fine-tuned with the full annotated dataset. 

We are motivated to study few-shot learning of language model with prompt-based language model fine-tuning. We exploit the rich information in unlabeled data with semi-supervised learning. Furthermore, we adopt an even smaller RoBERTa-base model as the backbone of our framework.

\begin{figure*}[ht]
    \centering
    \includegraphics[scale=0.7]{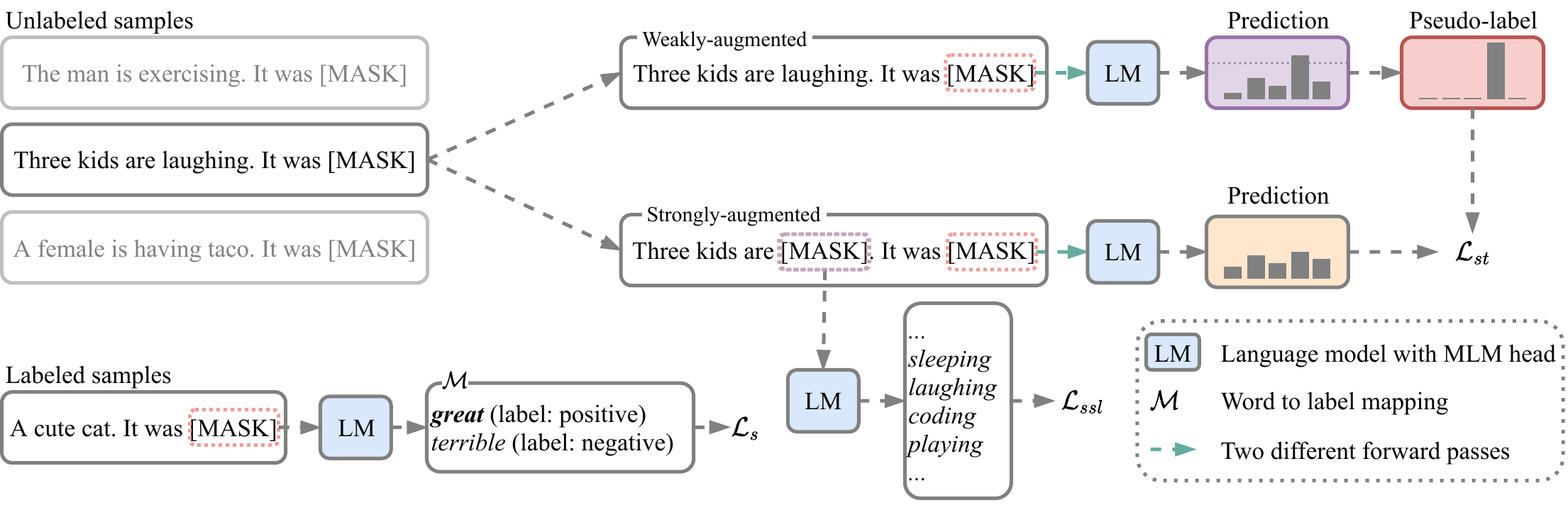}
    \vspace{-1mm}
    \caption{The learning process of SFLM on both labeled and unlabeled samples with three loss terms. For the supervised loss term $\mathcal{L}_s$, in SFLM, a pre-trained language model with a MLM head is used to get the predicted word from the template. Then the predicted word is mapped to the corresponding label with manually defined task-specific word to label mapping $\mathcal{M}$. Two loss terms are computed upon the unlabeled data: (1) We use masked language modeling to compute the self-supervised loss. (2) We use a weak augmented (dropout) sentence to get the pseudo-label, then force the prediction given by a strongly-augmented (random mask) view against the pseudo label via the self-training loss.} 
    \vspace{-5mm}
    \label{fig:framework}
\end{figure*}

\subsection{Self-Training}
\label{self-training-rw}
Self-training refers to the process of creating pseudo-labels on unlabeled data with a pre-trained teacher model,  then applying these labeled data to train a student model. It is a simple and effective semi-supervised approach, which has benefited a wide range of tasks, such as image classification~\citep{Xie2020SelfTrainingWN}, neural sequence generation~\citep{He2020RevisitingSF}, and parsing~\citep{mcclosky-etal-2006-effective}. Generally, sophisticated learning algorithms~\citep{Sohn2020FixMatchSS}, and a large corpus of task-relevant data~\citep{Xie2020UnsupervisedDA} are required for self-training to work well. 

From the algorithm perspective, FixMatch~\citep{Sohn2020FixMatchSS} is a simple and effective self-training framework for image classification, which unifies consistency regularization and pseudo-labeling. In our work, we transfer this useful framework to language model few-shot learning by exploring various text augmentation techniques for fine-tuning the pre-trained language model.

From the data perspective, several recent works have shown the effectiveness of self-training for language model fine-tuning~\citep{Du2020SelftrainingIP,schick2020s,schick-schutze-2021-exploiting} leveraging a large amount of unlabeled data. PET~\citep{schick2020s} adopts prompt-based fine-tuning and self-training for language model few-shot learning. This approach assumes the presence of a large number of unlabeled in-domain data (roughly 10,000 examples per class). In addition, \citet{Du2020SelftrainingIP} propose to retrieve task-relevant unlabeled data from a large-scale open-domain sentence bank. A paraphrase-based universal sentence encoder is designed to output sentence-level vectors for computing cosine similarity between labeled sentences and unlabeled ones in the sentence bank.

Unlike the prior studies, which rely on a large amount of unlabeled data and expensive computation resources, we do not assume the availability of abundant in-domain unlabeled data. Instead, we tackle the in-domain data constraint via improving data efficiency, i.e., proposing a scalable and effective self-training framework leveraging only a few unlabeled data.

\section{Methodology}
\textbf{Problem setup:} Our goal is to adapt pre-trained language models to downstream tasks in a few-shot setting. The model, $m$ should correctly classify unseen examples leveraging very few labeled data points from each class. Let $\mathcal{X}$ denote a small set of labeled training data with $N$ samples per class and an unlabeled dataset, $\mathcal{U}$ from the same task domain as $\mathcal{X}$. Assume that this unlabeled dataset has very limited size $\mu N$ per class, where $\mu$ is the ratio between the size of $\mathcal{X}$ and that of $\mathcal{U}$. 

During training, let each batch consist of $B$ labeled data points, $\mathcal{X_B}$, and $\mu B$ unlabeled data points, $\mathcal{U_B}$:

\begin{eqnarray}
    \mathcal{X_B}=\{(x_i,y_i):i\in(1,...,B)\} \\
    \mathcal{U_B}=\{u_i:i\in (1,...,\mu B)\}
\end{eqnarray}

Figure~\ref{fig:framework} illustrates the learning process with an example, containing one labeled and three unlabeled data samples. SFLM is optimized with following loss function:
\begin{equation}
    \label{main_loss}
    \mathcal{L}=\mathcal{L}_s + \lambda_1\mathcal{L}_{st}+\lambda_2\mathcal{L}_{ssl}
\end{equation}
where $\mathcal{L}_s$ is the prompt-based supervised loss applied to the labeled data~\citep{gao2020making}, $\mathcal{L}_{st}$ and $\mathcal{L}_{ssl}$ refer to self-training loss and self-supervised loss applied to the unlabeled data accordingly, while $\lambda_1$ and $\lambda_2$ are fixed scalar hyper-parameters controlling the relative weight of the unlabeled loss terms.

\textbf{Prompt-based supervised loss:} The prompt-based supervised loss is motivated by LM-BFF~\citep{gao2020making}. The classification is re-formulated as a language modeling task, in which the probability of class prediction $y_i\in\mathcal{Y}$ is,
\begin{equation}
    p_m(y_i|x_i)=p_m([\mathrm{MASK}]=\mathcal{M^{'}}(y_{i}|x^{prompt}_{i})
\end{equation}
where $\mathcal{M^{'}}$ refers a mapping from task labels to the corresponding words\footnote{$\mathcal{M^{'}}$ is the inverse operation of $\mathcal{M}$ as shown in figure~\ref{fig:framework}}, and $x^{prompt}_{i}$ is the reconstructed input sentence with task-specific template. For instance, in a sentence-level binary classification task, the input sentence $x_i$ is reconstructed as:
\begin{equation}
    x^{prompt}_{i}=x_i \circ \mathrm{It\ was\ [MASK].}
\end{equation}
where $\circ$ denotes the string concatenation operation.

Instead of using an additional classifier, the pre-trained masked language modeling head decides which word to be filled in the masked position. Then we could fine-tune the model with the standard cross-entropy loss:
\begin{equation}
    \mathcal{L}_s=\frac{1}{B}\sum_{i=1}^{B}H(y_i,p_m(y_i|x_i))
\end{equation}

\textbf{Self-training loss:} For each unlabeled sentence $u_i$, we obtain the weakly-augmented version $\alpha(u_i)$ and the strongly-augmented version $\mathcal{A}(u_i)$, where $\alpha$ and $\mathcal{A}$ refers to different augmentation strategies. The self-training process consists of two stages. Firstly, we assign a pseudo label to each unlabeled sentence in the batch by computing the output probability distribution corresponding to the weakly-augmented input sentence $\alpha(u_i)$, defined as $q_i=p_m(y_i|\alpha(u_i))$. The pseudo label, $\hat{q}_i$, is obtained by $\hat{q}_i=\arg\max{(q_i)}$. Secondly, we compute the prompt-based cross-entropy loss between $\hat{q}_i$ and the prediction corresponding to the strongly-augmented input sentence $\mathcal{A}(u_i)$. The self-training loss is defined as, 
\begin{equation}
\begin{split}
    \mathcal{L}_{st}= & \frac{1}{\mu B}\sum_{i=1}^{\mu B}\mathbbm{1}(max(q_i)\geq \tau)\ \cdot \\
    & \mathrm{H}(\hat{q}_i,p_m(y_i|\mathcal{A}(u_i)))
\end{split}
\end{equation}
where $\tau$ defines the threshold above which we retain a pseudo-label.

\citet{Sohn2020FixMatchSS} adopt AutoAugment~\citep{Cubuk2018AutoAugmentLA} for image augmentation, and highlight the importance of applying proper augmentation techniques in self-training. Text augmentation techniques can be tricky due to the discrete nature of text data. The recent successes in representation learning~\citep{devlin-etal-2019-bert,Gao2021SimCSESC} motivate us to purely rely on dropout for our weak augmentation, and random token masking for our strong augmentation. 

Specifically, the surface forms of weakly-augmented sentences remain unchanged: $\alpha(u_i) = u_i$. For strong augmentation, we randomly replace 15\% of the tokens in $\mathcal{A}(u_i)$ with the special mask token, $[\mathrm{MASK}]$. Then, we input $\alpha(u_i)$ and $\mathcal{A}(u_i)$ to the language model separately. Therefore, the two input sentences will undergo independent dropout operations (0.1 dropout rate by default), which can be considered as part of the data augmentation process (The green arrow in Figure~\ref{fig:framework}). We empirically show that the performance of our proposed augmentation techniques is superior against other common text augmentation techniques in Section~\ref{sec:experiment}.

\textbf{Self-supervised loss:} We also include an auxiliary self-supervised loss term, $\mathcal{L}_{ssl}$, for regularization purpose. The masked language model loss is used for its simplicity and efficiency.

\begin{table*}
\centering
\begin{tabular}{lcccccc}
\toprule
                   & \multicolumn{1}{c}{\begin{tabular}[c]{@{}c@{}}\textbf{SST-2}\\ (acc)\end{tabular}} & \multicolumn{1}{c}{\begin{tabular}[c]{@{}c@{}}\textbf{SST-5}\\ (acc)\end{tabular}}   & \multicolumn{1}{c}{\begin{tabular}[c]{@{}c@{}}\textbf{MR}\\ (acc)\end{tabular}}   & \multicolumn{1}{c}{\begin{tabular}[c]{@{}c@{}}\textbf{CR}\\ (acc)\end{tabular}}   & \multicolumn{1}{c}{\begin{tabular}[c]{@{}c@{}}\textbf{MPQA}\\ (acc)\end{tabular}} & \multicolumn{1}{c}{\begin{tabular}[c]{@{}c@{}}\textbf{Subj}\\ (acc)\end{tabular}} \\ \midrule
FT        & \multicolumn{1}{c}{78.3 (3.7)}                                & \multicolumn{1}{c}{36.4 (2.1)}                                                    & \multicolumn{1}{c}{69.5 (4.3)}                                                    & \multicolumn{1}{c}{78.6 (4.3)}                                                    & \multicolumn{1}{c}{69.2 (7.2)}                                                     & \multicolumn{1}{c}{89.6 (0.8)}                                                    \\ 
LM-BFF  & \multicolumn{1}{c}{89.9 (0.5)}   & \multicolumn{1}{c}{45.8 (3.1)} & \multicolumn{1}{c}{84.1 (1.7)}  & \multicolumn{1}{c}{89.5 (0.6)}    & \multicolumn{1}{c}{84.3 (1.1)}  & \multicolumn{1}{c}{88.3 (3.5)}            \\
PET-few$^\spadesuit$                & \multicolumn{1}{c}{89.8 (0.9)}     & \multicolumn{1}{c}{46.7 (0.8)}                                                & \multicolumn{1}{c}{84.2 (1.2)}                                                 & \multicolumn{1}{c}{89.4 (0.7)}                                                & \multicolumn{1}{c}{84.9 (0.9)}                                                & \multicolumn{1}{c}{90.0 (1.7)}  \\
PET-full$^\heartsuit$    & \multicolumn{1}{c}{90.2 (0.8)}    &  \multicolumn{1}{c}{46.0 (1.2)}                               & \multicolumn{1}{c}{85.0 (1.3)}                               & \multicolumn{1}{c}{88.9 (1.0)}                               & \multicolumn{1}{c}{84.1 (1.0)}                               & \multicolumn{1}{c}{91.4 (0.7)} \\ \midrule
SFLM               & \multicolumn{1}{c}{\textbf{91.0} (0.7)}          & \multicolumn{1}{c}{\textbf{47.7} (1.1)}                               & \multicolumn{1}{c}{\textbf{86.6} (0.4)}                               & \multicolumn{1}{c}{\textbf{90.8} (0.6)}                               & \multicolumn{1}{c}{\textbf{86.5} (0.3)}                               & \multicolumn{1}{c}{\textbf{92.4} (0.7)}                                         \\
\bottomrule
\toprule
                   & \multicolumn{1}{c}{\begin{tabular}[c]{@{}c@{}}\textbf{MNLI}\\ (acc)\end{tabular}}  & \multicolumn{1}{c}{\begin{tabular}[c]{@{}c@{}}\textbf{MNLI-mm}\\ (acc)\end{tabular}}  & \multicolumn{1}{c}{\begin{tabular}[c]{@{}c@{}}\textbf{SNLI}\\ (acc)\end{tabular}}  & \multicolumn{1}{c}{\begin{tabular}[c]{@{}c@{}}\textbf{QNLI}\\ (acc)\end{tabular}}  & \multicolumn{1}{c}{\begin{tabular}[c]{@{}c@{}}\textbf{RTE}\\ (acc)\end{tabular}}  & \multicolumn{1}{c}{\begin{tabular}[c]{@{}c@{}}\textbf{MRPC}\\ (F1)\end{tabular}}  \\ \midrule
FT        & \multicolumn{1}{c}{41.2 (2.2)}          & \multicolumn{1}{c}{42.9 (2.6)}                      & \multicolumn{1}{c}{43.9 (3.1)}                      & \multicolumn{1}{c}{60.3 (3.7)}                      & \multicolumn{1}{c}{50.3 (1.9)}                      & \multicolumn{1}{c}{70.8 (19.9)}         \\ 
LM-BFF & 60.2 (1.7) & 62.3 (1.4) & 65.8 (2.6) & 60.6 (2.1) & 66.2 (3.4) & 77.7 (0.8) \\
PET-few$^\spadesuit$         & \multicolumn{1}{c}{60.0 (1.8)}    & \multicolumn{1}{c}{61.6 (1.4)}  & \multicolumn{1}{c}{66.8 (2.8)} & \multicolumn{1}{c}{60.8 (2.5)} & \multicolumn{1}{c}{62.5 (1.7)} & \multicolumn{1}{c}{77.4 (5.0)}   \\
PET-full$^\heartsuit$                                     & \multicolumn{1}{c}{\textbf{62.6} (2.9)}                                          & \multicolumn{1}{c}{\textbf{64.8} (2.2)}                                          & \multicolumn{1}{c}{\textbf{67.8} (3.5)}                                         & \multicolumn{1}{c}{\textbf{61.3} (4.0)}                                         & \multicolumn{1}{c}{65.5 (2.3)}                                         & \multicolumn{1}{c}{77.5 (4.5)}                                             \\ \midrule
SFLM               & \multicolumn{1}{c}{\textbf{62.6} (1.5)}                  & \multicolumn{1}{c}{64.7 (1.3)}                                         & \multicolumn{1}{c}{67.4 (2.7)}                                         & \multicolumn{1}{c}{61.0 (4.6)}                                         & \multicolumn{1}{c}{\textbf{67.3} (2.7)}                                         & \multicolumn{1}{c}{\textbf{81.8} (1.2)}                                         \\
\bottomrule
\end{tabular}

\caption{\label{main-result}
We use RoBERTa-base and report the average scores in all experiments, where \textit{acc} denotes the accuracy (\%), and \textit{F1} denotes the F1 score. The standard deviation is included in the bracket. We use $N=16$ (\# labeled examples per class), and $\mu=4$ (ratio of unlabeled data to labeled data) for few-shot experiments. Upper block shows the results on single sentence tasks, while lower block shows the results on sentence pair tasks. $\spadesuit$: we re-implement the PET~\citep{schick-schutze-2021-exploiting} based on LM-BFF~\citep{gao2020making}. $\heartsuit$: We treat the full traning set as the unlabeled dataset and fine-tune PET with it. The size of the full training set is kept to 10,000 samples for all tasks.} 
\end{table*}

\section{Experiment}

\label{sec:experiment}

\subsection{Setup}

We evaluate our model on two groups of tasks: (1) 6 standard single sentence classification tasks (SST-2~\citep{socher-etal-2013-recursive}, SST-5~\citep{socher-etal-2013-recursive}, MR~\citep{pang-lee-2005-seeing}, CR~\citep{hu2004mining}, MPQA~\citep{wiebe2005annotating}, Subj~\citep{pang-lee-2004-sentimental}) and (2) 6 sentence pair classification tasks (MNLI~\citep{williams-etal-2018-broad}, MNLI-mm~\citep{williams-etal-2018-broad}, SNLI~\citep{bowman-etal-2015-large}, QNLI~\citep{rajpurkar-etal-2016-squad}, RTE~\citep{10.1007/11736790_9, haim2006second, giampiccolo-etal-2007-third,bentivogli2009fifth}, MRPC~\citep{dolan-brockett-2005-automatically}). These tasks are adapted from the benchmarks in~\citep{conneau-kiela-2018-senteval, wang-etal-2018-glue}. 


We set $N$ to 16 and $\mu$ to 4. Following~\citep{gao2020making}. We randomly sample five different splits of ($\mathcal{X}_{train}$, $\mathcal{X}_{dev}$, $\mathcal{U}$) from the original training set. Five different models are trained with these splits. Then, we report the average performance of these five models on the original development set. As in the previous work~\citep{schick2020s}, the sampled unlabeled splits are carefully constructed to account for class balance. As few-shot learning can be unstable, and extremely sensitive to hyper-parameter selection, we also perform a grid search over several hyper-parameters (learning rate, batch size $B$, controlling weight of loss $\lambda$ and confidence threshold $\tau$) across different tasks. Finally, Adam~\citep{Kingma2015AdamAM} is used as the optimizer.

\subsection{Baselines}
We consider three baselines, namely  standard fine-tuning (\textbf{FT}), supervised learning~\citep{gao2020making} (\textbf{LM-BFF}),  semi-supervised learning~\citep{schick-schutze-2021-exploiting} (\textbf{PET}). We use RoBERTa-base~\citep{Liu2019RoBERTaAR}, which has 125M parameters, and the same task-specific manual prompt from~\citep{gao2020making}, including template and word-to-label mapping, for prompt-based fine-tuning.

\textbf{FT:} We directly fine-tune the language model with the sequence classification head on the few-shot labeled dataset.

\textbf{LM-BFF:} We choose current state-of-the-art LM-BFF~\citep{gao2020making} as our supervised baseline. We use the prompt with demonstration~\citep{gao2020making} implementation across all tasks for fair comparison. We retrain the model with the official code~\footnote{\url{https://github.com/princeton-nlp/LM-BFF}}.

\textbf{PET:} For fair comparison, We re-implement our own version of PET based on LM-BFF~\citep{gao2020making}, since LM-BFF largely benefits from prompt-based fine-tuning. Specifically, we remove the knowledge distillation on the standard sequence classifier in the original implementation. Instead, we fine-tune the prompt-based language model~\citep{gao2020making} with a mixed training set of labeled and pseudo-labeled data. We iteratively increase the amount of pseudo-labeled data in the training set for model fine-tuning. Through our extensive experiments, we find that our implementation outperforms the official implementation\footnote{\url{https://github.com/timoschick/pet}} across various tasks. In addition, we evaluate PET under two different settings: (1) using reduced unlabeled dataset, which is the same as our SFLM; (2) using the full training set for self-training, while we limit the number of unlabeled samples to 10,000 per class such that the amount of unlabeled samples across different tasks are kept in the same range. 

\subsection{Main Results}
Table~\ref{main-result} presents the performance of SFLM against baselines across various benchmarking tasks. Overall, our proposed SFLM consistently outperforms the supervised and semi-supervised methods by 2\% on average with the same amount of data, and by 1.2\% with 45 times less unlabeled data. Next, we summarize our observations over the experiment results.

First, we find that self-training can greatly improve the performance of vanilla prompt-based fine-tuning under few-shot setting, either using PET or SFLM. With a large amount of in-domain unlabeled data, even a simple iterative self-training approach can boost the performance by 3.13\% on Subj, and 0.88\% on average (PET-full vs. LM-BFF). This demonstrates the effectiveness of exploiting the rich information carried in the unlabeled data.

Second, in-domain unlabeled data is crucial to the success of semi-supervised methods. As we down-sample the unlabeled dataset size to 64 samples per class, The performance of PET barely has any improvement w.r.t LM-BFF. The performance even degrades in some tasks. For example, in tasks\footnote{The full unlabeled datasets for these tasks contain less than 5,000 sentences.} such as RTE, CR, and MRPC, PET doesn't perform as well as LM-BFF even using the full unlabeled dataset. 

Third, unlike PET-few, SFLM can still bring a significant improvement w.r.t LM-BFF even when the size of unlabeled data is limited. This observation implies that our approach utilizes the unlabeled data in a much more efficient way. SFLM also outperforms PET on 8 tasks out of 12. The exceptions are the four natural language inference tasks, of which the unlabeled datasets contain 10,000 unlabeled data samples. However, the performance difference between PET-full, which uses all the 10,000 unlabeled data samples, and SFLM in these four tasks is insignificant (less than 0.4\%). In addition, SFLM generally has lower variance compared to the baselines.

The major difference between SFLM and LM-BFF is that we utilize the rich information carried in the unlabeled data. The experiment results confirm our hypothesis about the usefulness of semi-supervised learning in language model few-shot learning. Furthermore, the major difference between SFLM and PET is the self-training algorithm. We include strong data augmentation technique for consistency regularization. This confirms that our proposed text augmentation techniques are crucial to the success of SFLM.

\subsection{Analysis of Data Efficiency}
\begin{figure}[!tb]
  \centering
    \subfigure{\includegraphics[width=0.49\linewidth]{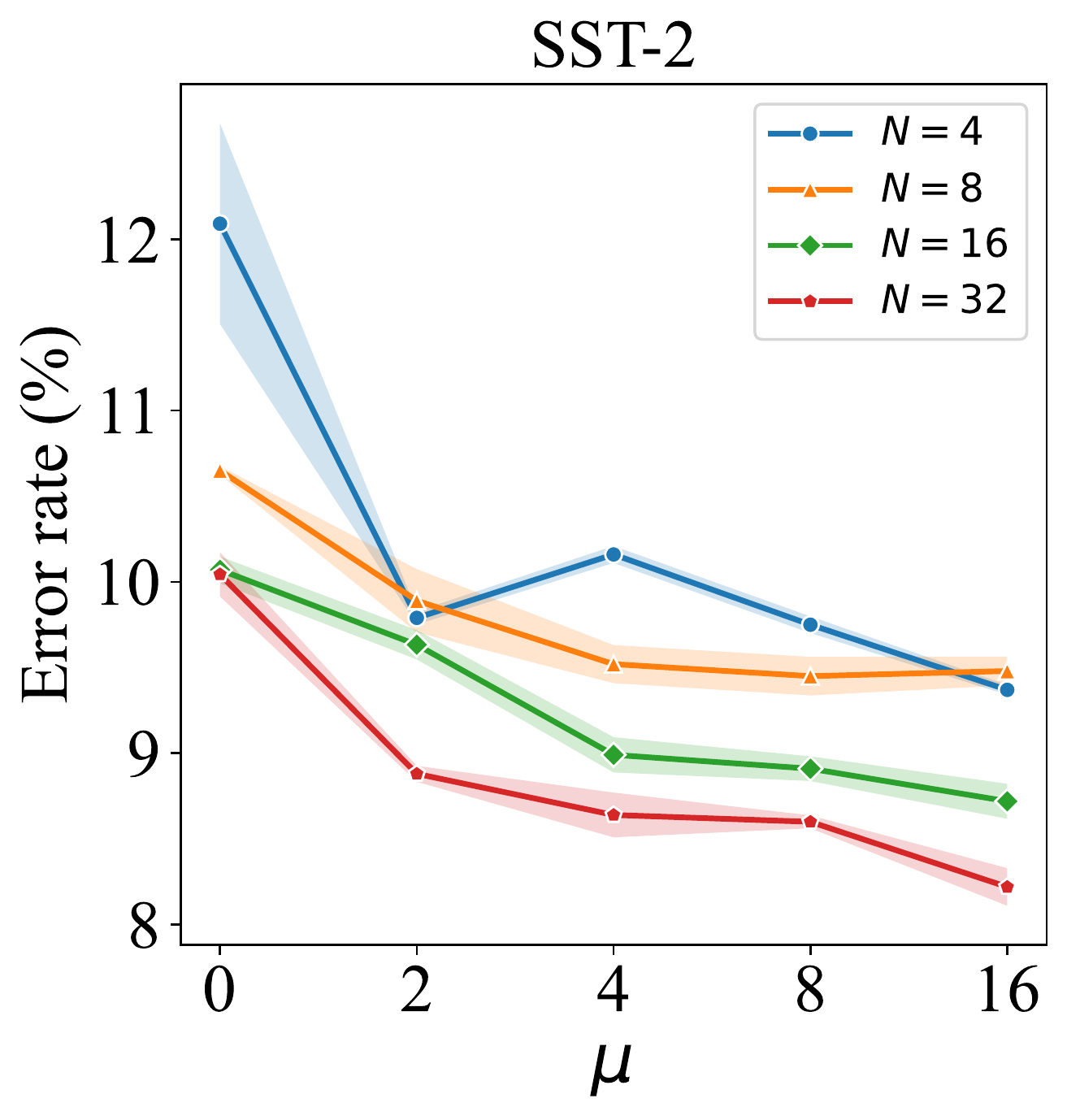}} 
	\subfigure{\includegraphics[width=0.49\linewidth]{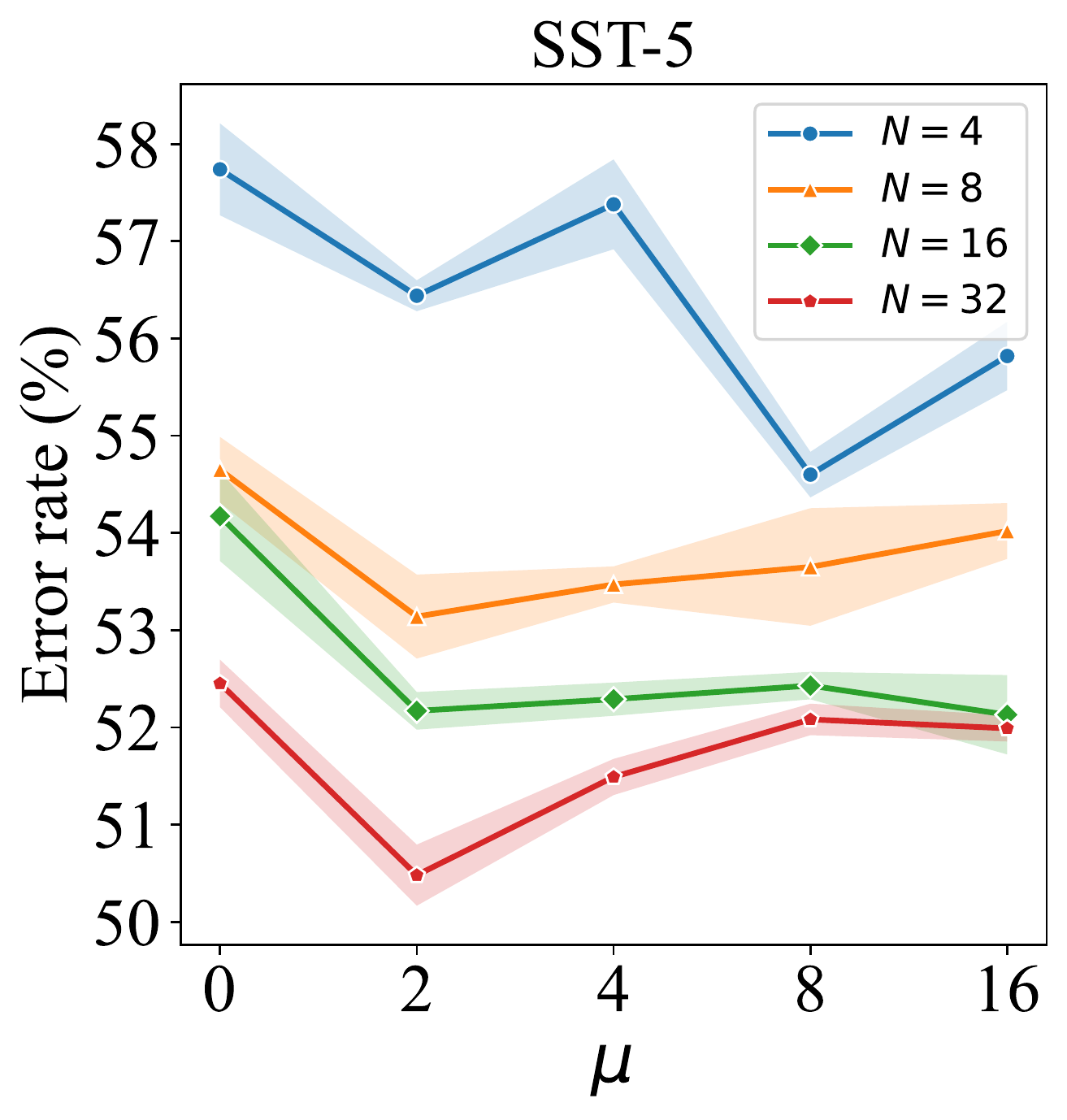}} \\
	\vspace{-3mm}
	\subfigure{\includegraphics[width=0.49\linewidth]{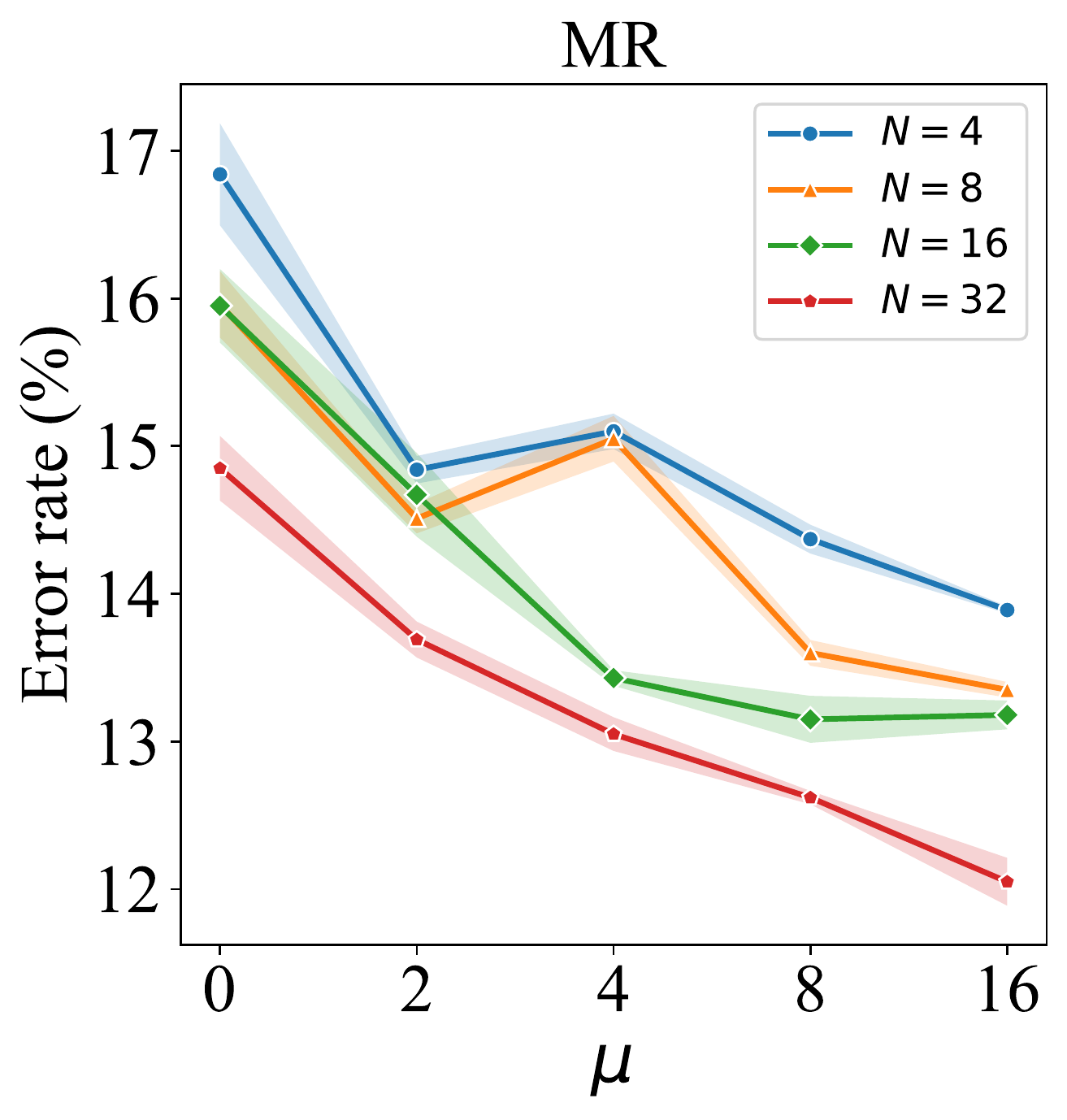}}
	\subfigure{\includegraphics[width=0.49\linewidth]{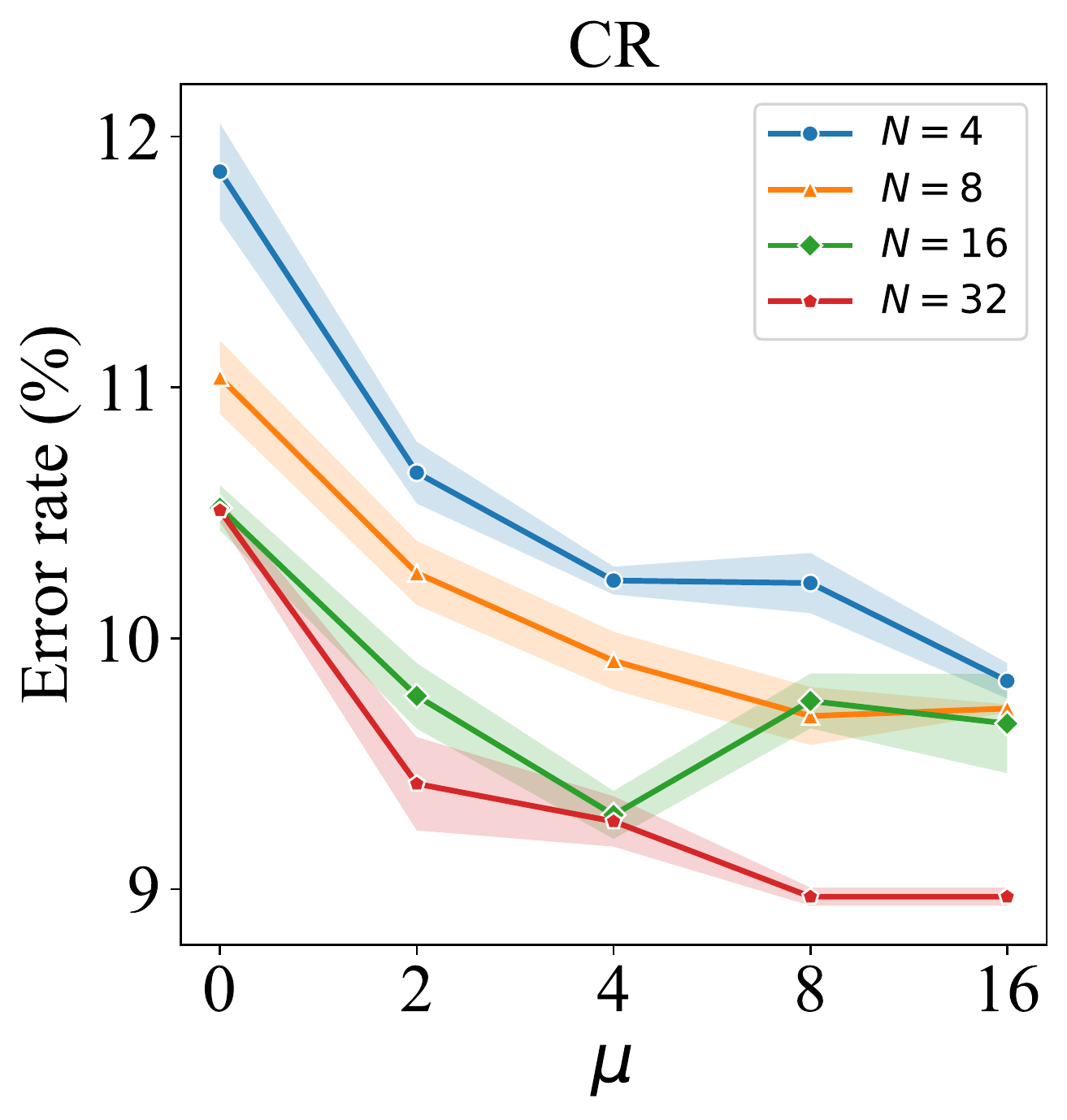}} \\
	\vspace{-3mm}
	\subfigure{\includegraphics[width=0.49\linewidth]{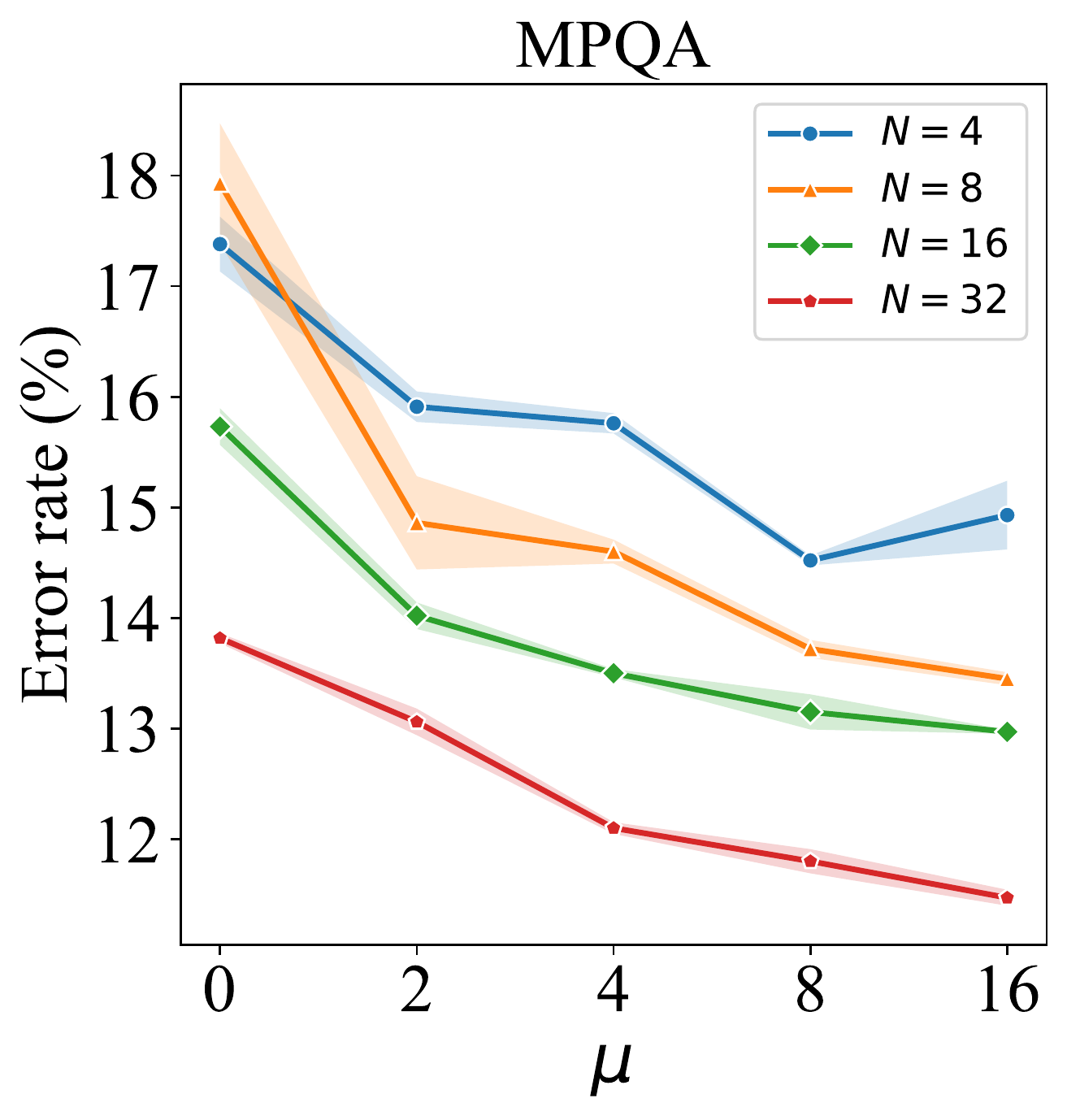}} 
	\subfigure{\includegraphics[width=0.49\linewidth]{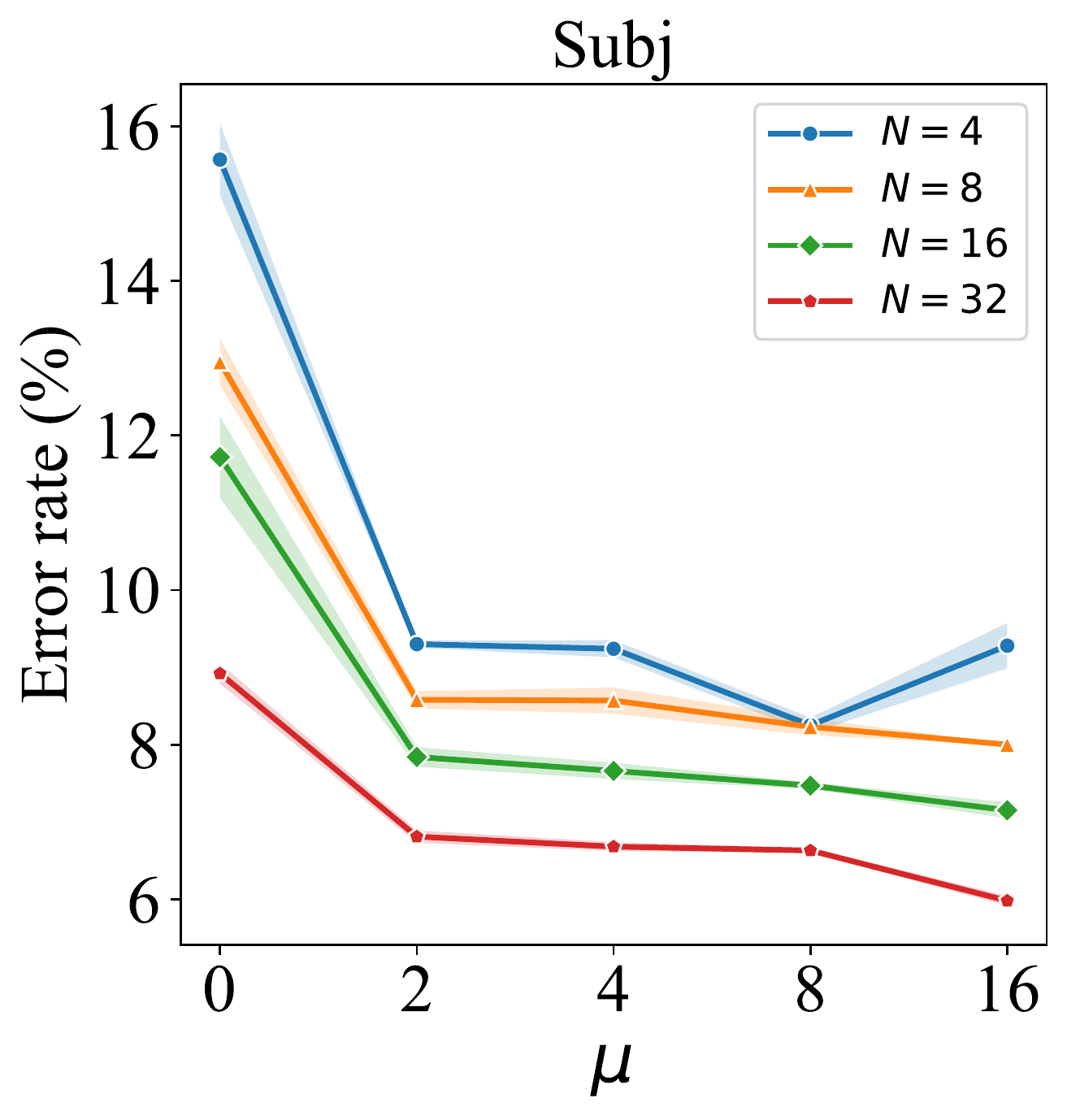}}
	\\
	\vspace{-3mm}
  \caption{Error rates of our SFLM under different $N$ (\# instances per class), and $\mu$ (ratio between unlabeled and labeled data)\protect\footnotemark. $\mu=0$ refers to vanilla LM-BFF.}
	\label{fig:data_efficiency}
\vspace{-5mm}
\end{figure}

One of the key questions in this study is how many labeled and unlabeled data are required. We provide an answer in Figure~\ref{fig:data_efficiency}, which illustrates the performance of SFLM with different combinations of $N$ and $\mu$. It can be observed that the error rate reduces generally as $\mu$ increases, that suggests SFLM benefits from more unlabeled data, with an exception in SST-5, where the best performance is achieved at $\mu=2$. a similar trend is also spotted for PET (see Table \ref{main-result}).

We note that SST-5 is the most difficult one among the six tasks. We observe that, with a relatively weak teacher model, self-training doesn't benefit from more unlabeled data. We speculate that by increasing unlabeled data, we introduce  more noise into the training process when we have a weak starting point.

In Figure~\ref{fig:data_efficiency}, we also observe that, for the simpler tasks, e.g., SST-2 and Subj, the gain obtained by more unlabeled data rapidly saturates when four times of unlabeled data are given, which is consistent with the finding in \citep{Sohn2020FixMatchSS}. We are encouraged to see that SFLM continues to improve as $\mu$ increases for the other three tasks and saturates later. In general, the trend of performance improvement by varying $\mu$ is consistent across different value of $N$. For instance, the performance gain is $\sim0.65\%$ for increasing $\mu$ from $2$ to $4$ across different $N$.

  \footnotetext{The performance in CR for ($N=32$, $\mu=16$) is put as the same as that of ($N=32$, $\mu=8$) due to limited amount of unlabeled data}

\begin{figure}[ht]
    \centering
    \includegraphics[scale=0.5]{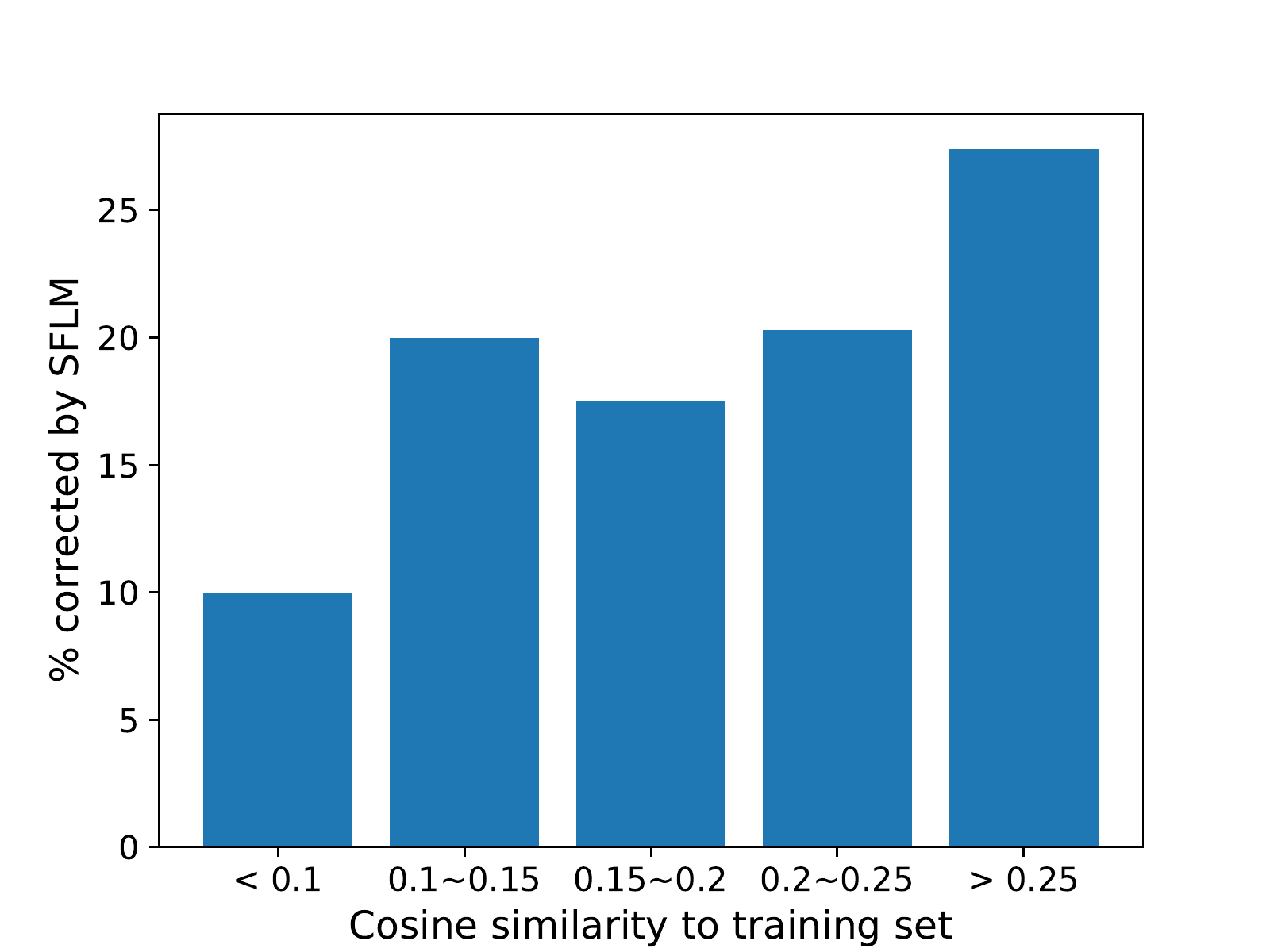}
    \caption{Percentage of examples with varying cosine similarities to training set corrected by SFLM.}
    \label{fig:analysis_st}
\end{figure}

To better understand what SLFM actually improves over the baselines, we further analyze the distribution of incorrectly-labelled examples corrected by SFLM, motivated by~\citep{wei2020theoretical}. First, we partition incorrectly-labelled examples into five bins based on the cosine similarity of their sentence embeddings given by SimCSE~\citep{Gao2021SimCSESC} w.r.t. the average embedding of the training set. Next, we show the percentage of examples in each bin whose labels are corrected by SFLM. As shown in Figure~\ref{fig:analysis_st}, SFLM is more likely to correct examples within the gold-labelled data’s neighborhood (the region surrounding a data point in the embedding space) than those away from the neighborhood.  In other words, the performance gain of SFLM is not only related to the unlabelled data size but also to the data distribution.

\begin{table*}[t]

\centering
\begin{tabular}{lccccccc}
\toprule
\textbf{Augmentation} & \textbf{SST-2} & \textbf{SST-5} & \textbf{MR} & \textbf{CR} & \textbf{MPQA} & \textbf{Subj} & \textbf{Avg} \\ 
\midrule
Dropout & 90.0 & 46.0 & 84.7 & 89.7 & 85.7 & 90.9 & 81.2 \\
Crop    & 89.6 & 45.7 & 83.6 & 88.4 & 85.2 & 88.8 & 80.2 \\
Swap    & \textbf{91.1} & 41.8 & 86.1 & 89.2 & 85.4 & 92.3 & 81.0    \\
Deletion& 90.3 & 44.1  & 83.7 & 89.8 & 86.3 & 90.0 & 80.7  \\ 
\midrule
Mask    & 91.0 & \textbf{47.7} & \textbf{86.6} & \textbf{90.8} & \textbf{86.5} & \textbf{92.4} & \textbf{82.5}   \\ 
\bottomrule
\end{tabular}

\caption{\label{aug-result}
Comparison of different data augmentation on six single sentence classification datasets (accuracy \%).} 
\end{table*}

SFLM is also scalable with different amounts of labeled data, and consistently outperforms LM-BFF by 3.2\% for $N=4$ and by 2\% for $N=32$ on average. Especially, SFLM exhibits significant improvement over LM-BFF under the extremely few-shot scenario. When $N=32$, the performance of LM-BFF saturates in simple tasks, e.g., SST-2, CR. However, SFLM continues to narrow the gap between few-shot learning and fine-tuning with the entire labeled dataset, which further validates the benefits of self-training to supervised learning.

\subsection{Augmentation Techniques}
It has been shown that strong data augmentation plays a crucial role in semi-supervised visual representation learning~\citep{Zoph2020RethinkingPA,Xie2020SelfTrainingWN,Sohn2020FixMatchSS}. The images can be augmented easily by cutout, flipping, and cropping~\citep{Zoph2020RethinkingPA}. However, very few works have been done on augmentation techniques for text~\citep{Xie2020SelfTrainingWN}\footnote{Recent works~\citep{Gao2021SimCSESC,Zhang2021BootstrappedUS,Wu2020CLEARCL}  discussed the effect of augmentation techniques on unsupervised sentence representation learning.  In SFLM, we focus on a few easy data augmentation strategies~\citep{wei-zou-2019-eda} for semi-supervised representation learning. }. 

Here, we study how different augmentation techniques would affect the model performance.  We fix \textit{Dropout} as the weak augmentation, and present the results of another three strong augmentation approaches, including \textit{Crop}, \textit{Swap} and \textit{Deletion} in Table~\ref{aug-result}. Specifically, \textit{Dropout} is same as the weak augmentation, we directly forward the original sentence into the language model. \textit{Crop} refers to randomly cropping the original sentence into a continuous span of 85\% of the original length. In terms of \textit{Swap}, we randomly swap two tokens in the sentence and repeat the same procedure 3 times. For \textit{Deletion}, we randomly delete 15\% of the tokens in a sentence. \textit{Mask} refers to randomly replacing 15\% of tokens in a sentence with the special [MASK] token. In this experiment, we keep $N=16$ and $\mu=4$.

We observe that the SFLM framework can also work with \textit{Dropout} and \textit{Swap}, as they still outperform PET on average by 0.4 and 0.2 respectively. However, they are less effective than \textit{Mask}. Another interesting finding is that \textit{Deletion}, which is similar to \textit{Mask}, yields poor  performance. We hypothesize that \textit{Deletion} and \textit{Crop} may adversely affect the original semantics, for example, deleting the word, \textit{not}, may reverse the meaning of the original sentence. In contrast, \textit{Mask} keeps the structure of sentences, and hence, it is easier to maintain the semantics by adding consistency regularization and MLM objectives. 

Furthermore, we empirically study the effect of different masking ratio on SST-2. 90.62\% of accuracy is obtained for 10\% masking, 90.14\% accuracy for 20\% masking, and the best performance of 91.0\% for 15\% masking. 

\begin{table}[h]

\centering
\begin{tabular}{llll}
\toprule
& \multicolumn{1}{c}{\textbf{CR}}  & \multicolumn{1}{c}{\textbf{MPQA}} & \multicolumn{1}{c}{\textbf{Subj}} \\
                   \midrule
\multicolumn{4}{c}{\emph{RoBERTa-base (\textbf{12-layers})}}\\
\midrule
FT     & \multicolumn{1}{c}{78.6}                         & \multicolumn{1}{c}{69.2}                                           & \multicolumn{1}{c}{89.6}  \\ 
LM-BFF  & \multicolumn{1}{c}{89.5}                                 & \multicolumn{1}{c}{84.3}                                           & \multicolumn{1}{c}{88.3}                \\
PET-few$^\spadesuit$       & \multicolumn{1}{c}{89.4}                                                & \multicolumn{1}{c}{84.9}                                                & \multicolumn{1}{c}{90.0}                                                    \\
PET-full$^\heartsuit$    & \multicolumn{1}{c}{88.9}                                                 & \multicolumn{1}{c}{84.1}                                                & \multicolumn{1}{c}{91.4}                                                 \\ \midrule
SFLM               & \multicolumn{1}{c}{90.8}                                         & \multicolumn{1}{c}{86.5}                                          & \multicolumn{1}{c}{92.4}                                              \\ \midrule
FT (full) & \multicolumn{1}{c}{89.6}                                                 & \multicolumn{1}{c}{87.4}                                                 & \multicolumn{1}{c}{96.9}                                  \\ 
                   \midrule
\multicolumn{4}{c}{\emph{DistilRoBERTa-base (\textbf{6-layers})}}\\
\midrule
FT        & \multicolumn{1}{c}{71.0}                   & \multicolumn{1}{c}{71.7}                                        &  \multicolumn{1}{c}{86.6}  \\ 
LM-BFF  & \multicolumn{1}{c}{84.8}    & \multicolumn{1}{c}{79.8}   & \multicolumn{1}{c}{87.5}                \\
PET-few$^\spadesuit$       & \multicolumn{1}{c}{86.5}      &    \multicolumn{1}{c}{80.6}                                              & \multicolumn{1}{c}{88.3}                                                    \\
PET-full$^\heartsuit$    & \multicolumn{1}{c}{87.2}    &  \multicolumn{1}{c}{80.6}                                                & \multicolumn{1}{c}{88.0}                                                 \\ \midrule
SFLM               & \multicolumn{1}{c}{87.9}                  & \multicolumn{1}{c}{82.0}                                       & \multicolumn{1}{c}{89.5}                                              \\ \midrule
FT (full) & \multicolumn{1}{c}{85.5}                           & \multicolumn{1}{c}{86.9}                                              & \multicolumn{1}{c}{96.6}                                  \\ 
\bottomrule
\end{tabular}

\caption{\label{scale-result}
Accuracy (\%) for systems with  language models of different size. We use $N=16$ (\# labeled examples per class), and $\mu=4$ (ratio of unlabeled data to labeled data) for few-shot experiments. $\spadesuit$, $\heartsuit$ follow the same definition in  Table \ref{main-result}. DistilRoBERa-base: 6-layer RoBERTa distilled from RoBERTa-base with 2 times speedup.} 
\end{table}
\subsection{Model Scale}
To put the SFLM framework under a stress test, we further apply SFLM on a smaller language model, distilled RoBERTa with 84M parameters, as reported in  Table~\ref{scale-result}. While DistilRoBERTa shows a similar performance as RoBERTa, it performs much worse under few-shot learning scenarios, e.g., LM-BFF has a significant performance drop of 5\% in CR and MPQA. We hypothesize that a robust language model of reasonable footprint is required for effective self-training. 

While SFLM consistently outperforms LM-BFF with a smaller language model, the overall performance of SFLM based on DistilRoBERTa-base also degrades sharply w.r.t that based on RoBERTa-base. This suggests the crucial role of the language model in our approach. 

Meanwhile, we find that PET better suits the small language model setting. For instance, the performance gain of PET-full w.r.t. LM-BFF increasing from 0.8\% to 1.1\% under the RoBERTa-base and DistilRoBERTa-base settings respectively. We hypothesize that more unlabeled data benefit few-shot learning of smaller language model. However, the performance of PET is still worse than that of SFLM under the DistilRoBERTa-base setting. 

Overall, the above observations confirm the effectiveness of our approach for language model few-shot learning regardless of the backbone language model.

\subsection{Zero-shot Transfer Across Tasks}
Lastly, we show that our proposed method can be easily extended to zero-shot transfer across tasks. Specifically, we assume that before few-shot learning on the target unlabeled dataset $\mathcal{U}$, the learner has access to an annotated dataset $\mathcal{D}$  known as the base dataset.\footnote{The concept of task transfer has been successfully applied to natural language understanding. For example, pre-training language model on MNLI before fine-tuning on RTE~\citep{Liu2019RoBERTaAR} yields better performance.} Accordingly,  we modify the learning objective as:
  \begin{equation}
  \label{eq:lpt}
      \mathcal{L} = \mathcal{L}_s^\mathcal{D}+\lambda_1\mathcal{L}_{st}^\mathcal{U}+\lambda_2\mathcal{L}_{ssl}^\mathcal{U}, 
  \end{equation}
where the last two terms are intended to encourage the model to capture knowledge specific to the target task.

We evaluate SFLM on three binary classification tasks, including sentiment analysis (MR and SST) and product reviews (CR). In the experiments, we use one of the three tasks as the source task and test on another two target tasks.  The SFLM model is based on RoBERTa-base and trained on 64 samples per class for $\mathcal{D}$ and $\mathcal{U}$, respectively. We compare SFLM with a baseline, which is a language model fine-tuned directly on the source dataset with a single prompt-based loss $\mathcal{L}_s^\mathcal{D}$.

The results are reported in Table~\ref{transfer-result}. SFLM, which adopts a small number of unlabeled target data, generally outperforms that with supervised fine-tuning  on source datasets. In particular, with MR as the source and CR as the target, SFLM obtains 1.3 points higher than the baseline model, which demonstrates that our proposed method has the flexibility to be applied to the cross-task zero-shot learning scenario.







\begin{table}
\centering
\begin{tabular}{lcccccc}
\toprule

 &  \multicolumn{3}{c}{{Transfer}} & \multicolumn{3}{c}{{ SFLM}} \\
\cmidrule(l{5pt}r{5pt}){2-4} \cmidrule(l{5pt}r{5pt}){5-7}
 &SST &MR  &CR  &SST &MR  &CR    \\
\midrule
SST &   -   &  87.3   & 89.7 & - & 87.3 &  90.8       \\
MR & 90.7 & - & 89.5 & 91.4 & - & 90.8   \\
CR & 89.7 & 84.5 & - & 90.3 & 85.5 & - \\ \bottomrule
\end{tabular}
\caption{\label{transfer-result}
Accuracy (\%) for zero-shot task transfer  between three sentiment classification tasks with RoBERTa-base. Transfer: a  language model fine-tuned on source dataset with a single prompt-based loss; We refer SST  to SST-2.} 
\end{table}
\section{Conclusion and Future Work}
In this paper, we present SFLM, a simple and effective self-training framework for few-shot learning of language model. Our approach addresses the few-shot language model fine-tuning problem with very limited labeled and unlabeled data with a self-training loss term unifying pseudo-labeling and consistency regularization. Through comprehensive experiments, we show that our approach outperforms previous state-of-the-art methods across tasks, data amount, and model scale. Despite its efficiency, SFLM also has several limitations. Compared to standard fine-tuning, SFLM requires more computational resources for unlabeled data. In addition, the performance gain by self-training is not proportional to the amount of unlabeled data. We leave it to future study. Namely, how to utilize large amount of unlabeled data efficiently through self-training.

\section*{Acknowledgements}

We would like to thank all the anonymous reviewers for their constructive comments. This work is partly supported by Human-Robot Interaction Phase 1 (Grant No. 19225 00054), National Research Foundation (NRF) Singapore under the National Robotics Programme; Human Robot Collaborative AI for AME (Grant No. A18A2b0046), NRF Singapore; National Natural Science Foundation of China (Grant NO. 61903178, 61906081, and U20A20306); Program for Guangdong Introducing Innovative and Entrepreneurial Teams (Grant No. 2017ZT03X386); Program for University Key Laboratory of Guangdong Province (Grant No. 2017KSYS008).

\bibliography{anthology,custom}

\begin{thebibliography}{52}
\expandafter\ifx\csname natexlab\endcsname\relax\def\natexlab#1{#1}\fi

\bibitem[{Bansal et~al.(2020)Bansal, Jha, Munkhdalai, and
  McCallum}]{bansal-etal-2020-self}
Trapit Bansal, Rishikesh Jha, Tsendsuren Munkhdalai, and Andrew McCallum. 2020.
\newblock \href {https://doi.org/10.18653/v1/2020.emnlp-main.38}
  {Self-supervised meta-learning for few-shot natural language classification
  tasks}.
\newblock In \emph{Proceedings of the 2020 Conference on Empirical Methods in
  Natural Language Processing (EMNLP)}, pages 522--534, Online. Association for
  Computational Linguistics.

\bibitem[{Bao et~al.(2020)Bao, Wu, Chang, and Barzilay}]{Bao2020FewshotTC}
Yujia Bao, Menghua Wu, Shiyu Chang, and Regina Barzilay. 2020.
\newblock Few-shot text classification with distributional signatures.
\newblock In \emph{Proc. of ICLR}.

\bibitem[{Bentivogli et~al.(2009)Bentivogli, Clark, Dagan, and
  Giampiccolo}]{bentivogli2009fifth}
Luisa Bentivogli, Peter Clark, Ido Dagan, and Danilo Giampiccolo. 2009.
\newblock The fifth pascal recognizing textual entailment challenge.
\newblock In \emph{TAC}.

\bibitem[{Blum and Mitchell(1998)}]{Blum1998CombiningLA}
Avrim Blum and Tom Mitchell. 1998.
\newblock Combining labeled and unlabeled data with co-training.
\newblock In \emph{Proc. of COLT}.

\bibitem[{Bowman et~al.(2015)Bowman, Angeli, Potts, and
  Manning}]{bowman-etal-2015-large}
Samuel~R. Bowman, Gabor Angeli, Christopher Potts, and Christopher~D. Manning.
  2015.
\newblock \href {https://doi.org/10.18653/v1/D15-1075} {A large annotated
  corpus for learning natural language inference}.
\newblock In \emph{Proceedings of the 2015 Conference on Empirical Methods in
  Natural Language Processing}, pages 632--642, Lisbon, Portugal. Association
  for Computational Linguistics.

\bibitem[{Brown et~al.(2020)Brown, Mann, Ryder, Subbiah, Kaplan, Dhariwal,
  Neelakantan, Shyam, Sastry, Askell, Agarwal, Herbert-Voss, Krueger, Henighan,
  Child, Ramesh, Ziegler, Wu, Winter, Hesse, Chen, Sigler, Litwin, Gray, Chess,
  Clark, Berner, McCandlish, Radford, Sutskever, and
  Amodei}]{Brown2020LanguageMA}
T.~Brown, Benjamin Mann, Nick Ryder, Melanie Subbiah, J.~Kaplan, Prafulla
  Dhariwal, Arvind Neelakantan, Pranav Shyam, Girish Sastry, Amanda Askell,
  Sandhini Agarwal, Ariel Herbert-Voss, Gretchen Krueger, T.~Henighan,
  R.~Child, A.~Ramesh, Daniel~M. Ziegler, Jeffrey Wu, Clemens Winter,
  Christopher Hesse, Mark Chen, Eric Sigler, Mateusz Litwin, Scott Gray,
  Benjamin Chess, J.~Clark, Christopher Berner, Sam McCandlish, A.~Radford,
  Ilya Sutskever, and Dario Amodei. 2020.
\newblock Language models are few-shot learners.
\newblock \emph{Proc. of NeurIPS}.

\bibitem[{Clark et~al.(2020)Clark, Luong, Le, and Manning}]{clark2020electra}
Kevin Clark, Minh-Thang Luong, Quoc~V Le, and Christopher~D Manning. 2020.
\newblock Electra: Pre-training text encoders as discriminators rather than
  generators.
\newblock In \emph{Proc. of ICLR}.

\bibitem[{Conneau and Kiela(2018)}]{conneau-kiela-2018-senteval}
Alexis Conneau and Douwe Kiela. 2018.
\newblock \href {https://www.aclweb.org/anthology/L18-1269} {{S}ent{E}val: An
  evaluation toolkit for universal sentence representations}.
\newblock In \emph{Proceedings of the Eleventh International Conference on
  Language Resources and Evaluation ({LREC} 2018)}, Miyazaki, Japan. European
  Language Resources Association (ELRA).

\bibitem[{Cubuk et~al.(2018)Cubuk, Zoph, Man{\'e}, Vasudevan, and
  Le}]{Cubuk2018AutoAugmentLA}
E.~D. Cubuk, Barret Zoph, Dandelion Man{\'e}, Vijay Vasudevan, and Quoc~V. Le.
  2018.
\newblock Autoaugment: Learning augmentation policies from data.
\newblock \emph{ArXiv}.

\bibitem[{Dagan et~al.(2005)Dagan, Glickman, and Magnini}]{10.1007/11736790_9}
Ido Dagan, Oren Glickman, and Bernardo Magnini. 2005.
\newblock \href {https://doi.org/10.1007/11736790_9} {The pascal recognising
  textual entailment challenge}.
\newblock page 177–190.

\bibitem[{Devlin et~al.(2019)Devlin, Chang, Lee, and
  Toutanova}]{devlin-etal-2019-bert}
Jacob Devlin, Ming-Wei Chang, Kenton Lee, and Kristina Toutanova. 2019.
\newblock \href {https://doi.org/10.18653/v1/N19-1423} {{BERT}: Pre-training of
  deep bidirectional transformers for language understanding}.
\newblock In \emph{Proceedings of the 2019 Conference of the North {A}merican
  Chapter of the Association for Computational Linguistics: Human Language
  Technologies, Volume 1 (Long and Short Papers)}, pages 4171--4186,
  Minneapolis, Minnesota. Association for Computational Linguistics.

\bibitem[{Dolan and Brockett(2005)}]{dolan-brockett-2005-automatically}
William~B. Dolan and Chris Brockett. 2005.
\newblock \href {https://www.aclweb.org/anthology/I05-5002} {Automatically
  constructing a corpus of sentential paraphrases}.
\newblock In \emph{Proceedings of the Third International Workshop on
  Paraphrasing ({IWP}2005)}.

\bibitem[{Du et~al.(2020)Du, Grave, Gunel, Chaudhary, Celebi, Auli, Stoyanov,
  and Conneau}]{Du2020SelftrainingIP}
Jingfei Du, E.~Grave, Beliz Gunel, Vishrav Chaudhary, Onur Celebi, Michael
  Auli, Ves Stoyanov, and Alexis Conneau. 2020.
\newblock Self-training improves pre-training for natural language
  understanding.
\newblock In \emph{ArXiv}.

\bibitem[{Finn et~al.(2017)Finn, Abbeel, and Levine}]{finn2017model}
Chelsea Finn, Pieter Abbeel, and Sergey Levine. 2017.
\newblock Model-agnostic meta-learning for fast adaptation of deep networks.
\newblock In \emph{Proc. of ICML}.

\bibitem[{Gao et~al.(2020)Gao, Fisch, and Chen}]{gao2020making}
Tianyu Gao, Adam Fisch, and Danqi Chen. 2020.
\newblock Making pre-trained language models better few-shot learners.
\newblock \emph{arXiv preprint arXiv:2012.15723}.

\bibitem[{Gao et~al.(2021)Gao, Yao, and Chen}]{Gao2021SimCSESC}
Tianyu Gao, Xingcheng Yao, and Danqi Chen. 2021.
\newblock Simcse: Simple contrastive learning of sentence embeddings.
\newblock \emph{ArXiv}, abs/2104.08821.

\bibitem[{Giampiccolo et~al.(2007)Giampiccolo, Magnini, Dagan, and
  Dolan}]{giampiccolo-etal-2007-third}
Danilo Giampiccolo, Bernardo Magnini, Ido Dagan, and Bill Dolan. 2007.
\newblock \href {https://www.aclweb.org/anthology/W07-1401} {The third {PASCAL}
  recognizing textual entailment challenge}.
\newblock In \emph{Proceedings of the {ACL}-{PASCAL} Workshop on Textual
  Entailment and Paraphrasing}, pages 1--9, Prague. Association for
  Computational Linguistics.

\bibitem[{Gunel et~al.(2020)Gunel, Du, Conneau, and
  Stoyanov}]{Gunel2020SupervisedCL}
Beliz Gunel, Jingfei Du, Alexis Conneau, and Ves Stoyanov. 2020.
\newblock Supervised contrastive learning for pre-trained language model
  fine-tuning.
\newblock \emph{ArXiv}, abs/2011.01403.

\bibitem[{Haim et~al.(2006)Haim, Dagan, Dolan, Ferro, Giampiccolo, Magnini, and
  Szpektor}]{haim2006second}
R~Bar Haim, Ido Dagan, Bill Dolan, Lisa Ferro, Danilo Giampiccolo, Bernardo
  Magnini, and Idan Szpektor. 2006.
\newblock The second pascal recognising textual entailment challenge.
\newblock In \emph{Proc. of the Second PASCAL Challenges Workshop on
  Recognising Textual Entailment}.

\bibitem[{Han et~al.(2018)Han, Zhu, Yu, Wang, Yao, Liu, and
  Sun}]{han-etal-2018-fewrel}
Xu~Han, Hao Zhu, Pengfei Yu, Ziyun Wang, Yuan Yao, Zhiyuan Liu, and Maosong
  Sun. 2018.
\newblock \href {https://doi.org/10.18653/v1/D18-1514} {{F}ew{R}el: A
  large-scale supervised few-shot relation classification dataset with
  state-of-the-art evaluation}.
\newblock In \emph{Proceedings of the 2018 Conference on Empirical Methods in
  Natural Language Processing}, pages 4803--4809, Brussels, Belgium.
  Association for Computational Linguistics.

\bibitem[{He et~al.(2019)He, Gu, Shen, and Ranzato}]{He2020RevisitingSF}
Junxian He, Jiatao Gu, Jiajun Shen, and Marc'Aurelio Ranzato. 2019.
\newblock Revisiting self-training for neural sequence generation.
\newblock In \emph{Proc. of ICLR}.

\bibitem[{Hu and Liu(2004)}]{hu2004mining}
Minqing Hu and Bing Liu. 2004.
\newblock Mining and summarizing customer reviews.
\newblock In \emph{Proceedings of the tenth ACM SIGKDD international conference
  on Knowledge discovery and data mining}.

\bibitem[{Kingma and Ba(2015)}]{Kingma2015AdamAM}
Diederik~P. Kingma and Jimmy Ba. 2015.
\newblock Adam: A method for stochastic optimization.
\newblock In \emph{Proc. of ICLR}.

\bibitem[{Lan et~al.(2020)Lan, Chen, Goodman, Gimpel, Sharma, and
  Soricut}]{Lan2020ALBERTAL}
Zhenzhong Lan, Mingda Chen, Sebastian Goodman, Kevin Gimpel, Piyush Sharma, and
  Radu Soricut. 2020.
\newblock Albert: A lite bert for self-supervised learning of language
  representations.
\newblock In \emph{Proc. of ICLR}.

\bibitem[{Liu et~al.(2019)Liu, Ott, Goyal, Du, Joshi, Chen, Levy, Lewis,
  Zettlemoyer, and Stoyanov}]{Liu2019RoBERTaAR}
Y.~Liu, Myle Ott, Naman Goyal, Jingfei Du, Mandar Joshi, Danqi Chen, Omer Levy,
  M.~Lewis, Luke Zettlemoyer, and Veselin Stoyanov. 2019.
\newblock Roberta: A robustly optimized bert pretraining approach.
\newblock In \emph{ArXiv}.

\bibitem[{McClosky et~al.(2006)McClosky, Charniak, and
  Johnson}]{mcclosky-etal-2006-effective}
David McClosky, Eugene Charniak, and Mark Johnson. 2006.
\newblock \href {https://www.aclweb.org/anthology/N06-1020} {Effective
  self-training for parsing}.
\newblock In \emph{Proceedings of the Human Language Technology Conference of
  the {NAACL}, Main Conference}, pages 152--159, New York City, USA.
  Association for Computational Linguistics.

\bibitem[{Pang and Lee(2004)}]{pang-lee-2004-sentimental}
Bo~Pang and Lillian Lee. 2004.
\newblock \href {https://doi.org/10.3115/1218955.1218990} {A sentimental
  education: Sentiment analysis using subjectivity summarization based on
  minimum cuts}.
\newblock In \emph{Proceedings of the 42nd Annual Meeting of the Association
  for Computational Linguistics ({ACL}-04)}, pages 271--278, Barcelona, Spain.

\bibitem[{Pang and Lee(2005)}]{pang-lee-2005-seeing}
Bo~Pang and Lillian Lee. 2005.
\newblock \href {https://doi.org/10.3115/1219840.1219855} {Seeing stars:
  Exploiting class relationships for sentiment categorization with respect to
  rating scales}.
\newblock In \emph{Proceedings of the 43rd Annual Meeting of the Association
  for Computational Linguistics ({ACL}{'}05)}, pages 115--124, Ann Arbor,
  Michigan. Association for Computational Linguistics.

\bibitem[{Qiu et~al.(2019)Qiu, Cho, Ma, and Campbell}]{qiu-etal-2019-graph}
Zimeng Qiu, Eunah Cho, Xiaochun Ma, and William Campbell. 2019.
\newblock \href {https://doi.org/10.18653/v1/D19-5318} {Graph-based
  semi-supervised learning for natural language understanding}.
\newblock In \emph{Proceedings of the Thirteenth Workshop on Graph-Based
  Methods for Natural Language Processing (TextGraphs-13)}, pages 151--158,
  Hong Kong. Association for Computational Linguistics.

\bibitem[{Radford et~al.(2019)Radford, Wu, Child, Luan, Amodei, and
  Sutskever}]{radford2019language}
Alec Radford, Jeffrey Wu, Rewon Child, David Luan, Dario Amodei, and Ilya
  Sutskever. 2019.
\newblock Language models are unsupervised multitask learners.
\newblock \emph{OpenAI blog}.

\bibitem[{Raffel et~al.(2020)Raffel, Shazeer, Roberts, Lee, Narang, Matena,
  Zhou, Li, and Liu}]{Raffel2020ExploringTL}
Colin Raffel, Noam~M. Shazeer, Adam Roberts, Katherine Lee, Sharan Narang,
  Michael Matena, Yanqi Zhou, W.~Li, and Peter~J. Liu. 2020.
\newblock Exploring the limits of transfer learning with a unified text-to-text
  transformer.
\newblock \emph{J. Mach. Learn. Res.}, 21:140:1--140:67.

\bibitem[{Rajpurkar et~al.(2016)Rajpurkar, Zhang, Lopyrev, and
  Liang}]{rajpurkar-etal-2016-squad}
Pranav Rajpurkar, Jian Zhang, Konstantin Lopyrev, and Percy Liang. 2016.
\newblock \href {https://doi.org/10.18653/v1/D16-1264} {{SQ}u{AD}: 100,000+
  questions for machine comprehension of text}.
\newblock In \emph{Proceedings of the 2016 Conference on Empirical Methods in
  Natural Language Processing}, pages 2383--2392, Austin, Texas. Association
  for Computational Linguistics.

\bibitem[{Schick and Sch{\"u}tze(2020)}]{schick2020s}
Timo Schick and Hinrich Sch{\"u}tze. 2020.
\newblock It's not just size that matters: Small language models are also
  few-shot learners.
\newblock In \emph{Proc. of NAACL}.

\bibitem[{Schick and Sch{\"u}tze(2021)}]{schick-schutze-2021-exploiting}
Timo Schick and Hinrich Sch{\"u}tze. 2021.
\newblock \href {https://www.aclweb.org/anthology/2021.eacl-main.20}
  {Exploiting cloze-questions for few-shot text classification and natural
  language inference}.
\newblock In \emph{Proceedings of the 16th Conference of the European Chapter
  of the Association for Computational Linguistics: Main Volume}, pages
  255--269, Online. Association for Computational Linguistics.

\bibitem[{Snell et~al.(2017)Snell, Swersky, and Zemel}]{snell2017prototypical}
Jake Snell, Kevin Swersky, and Richard Zemel. 2017.
\newblock Prototypical networks for few-shot learning.
\newblock In \emph{Proc. of NeurlPS}.

\bibitem[{Socher et~al.(2013)Socher, Perelygin, Wu, Chuang, Manning, Ng, and
  Potts}]{socher-etal-2013-recursive}
Richard Socher, Alex Perelygin, Jean Wu, Jason Chuang, Christopher~D. Manning,
  Andrew Ng, and Christopher Potts. 2013.
\newblock \href {https://www.aclweb.org/anthology/D13-1170} {Recursive deep
  models for semantic compositionality over a sentiment treebank}.
\newblock In \emph{Proceedings of the 2013 Conference on Empirical Methods in
  Natural Language Processing}, pages 1631--1642, Seattle, Washington, USA.
  Association for Computational Linguistics.

\bibitem[{Sohn et~al.(2020)Sohn, Berthelot, Carlini, Zhang, Zhang, Raffel,
  Cubuk, Kurakin, and Li}]{Sohn2020FixMatchSS}
Kihyuk Sohn, David Berthelot, Nicholas Carlini, Zizhao Zhang, Han Zhang,
  Colin~A Raffel, Ekin~Dogus Cubuk, Alexey Kurakin, and Chun-Liang Li. 2020.
\newblock Fixmatch: Simplifying semi-supervised learning with consistency and
  confidence.
\newblock In \emph{Proc. of NeurIPS}.

\bibitem[{Vinyals et~al.(2016)Vinyals, Blundell, Lillicrap, Kavukcuoglu, and
  Wierstra}]{vinyals2016matching}
Oriol Vinyals, Charles Blundell, Timothy Lillicrap, Koray Kavukcuoglu, and Daan
  Wierstra. 2016.
\newblock Matching networks for one shot learning.
\newblock In \emph{Proc. of NeurlPS}.

\bibitem[{Wang et~al.(2018)Wang, Singh, Michael, Hill, Levy, and
  Bowman}]{wang-etal-2018-glue}
Alex Wang, Amanpreet Singh, Julian Michael, Felix Hill, Omer Levy, and Samuel
  Bowman. 2018.
\newblock \href {https://doi.org/10.18653/v1/W18-5446} {{GLUE}: A multi-task
  benchmark and analysis platform for natural language understanding}.
\newblock In \emph{Proceedings of the 2018 {EMNLP} Workshop {B}lackbox{NLP}:
  Analyzing and Interpreting Neural Networks for {NLP}}, pages 353--355,
  Brussels, Belgium. Association for Computational Linguistics.

\bibitem[{Wei et~al.(2020)Wei, Shen, Chen, and Ma}]{wei2020theoretical}
Colin Wei, Kendrick Shen, Yining Chen, and Tengyu Ma. 2020.
\newblock Theoretical analysis of self-training with deep networks on unlabeled
  data.
\newblock In \emph{Proc. of ICLR}.

\bibitem[{Wei and Zou(2019)}]{wei-zou-2019-eda}
Jason Wei and Kai Zou. 2019.
\newblock \href {https://doi.org/10.18653/v1/D19-1670} {{EDA}: Easy data
  augmentation techniques for boosting performance on text classification
  tasks}.
\newblock In \emph{Proceedings of the 2019 Conference on Empirical Methods in
  Natural Language Processing and the 9th International Joint Conference on
  Natural Language Processing (EMNLP-IJCNLP)}, pages 6382--6388, Hong Kong,
  China. Association for Computational Linguistics.

\bibitem[{Wiebe et~al.(2005)Wiebe, Wilson, and Cardie}]{wiebe2005annotating}
Janyce Wiebe, Theresa Wilson, and Claire Cardie. 2005.
\newblock Annotating expressions of opinions and emotions in language.
\newblock \emph{Language resources and evaluation}, 39(2):165--210.

\bibitem[{Williams et~al.(2018)Williams, Nangia, and
  Bowman}]{williams-etal-2018-broad}
Adina Williams, Nikita Nangia, and Samuel Bowman. 2018.
\newblock \href {https://doi.org/10.18653/v1/N18-1101} {A broad-coverage
  challenge corpus for sentence understanding through inference}.
\newblock In \emph{Proceedings of the 2018 Conference of the North {A}merican
  Chapter of the Association for Computational Linguistics: Human Language
  Technologies, Volume 1 (Long Papers)}, pages 1112--1122, New Orleans,
  Louisiana. Association for Computational Linguistics.

\bibitem[{Wu et~al.(2020)Wu, Wang, Gu, Khabsa, Sun, and Ma}]{Wu2020CLEARCL}
Zhuofeng Wu, Sinong Wang, Jiatao Gu, Madian Khabsa, Fei Sun, and Hao Ma. 2020.
\newblock Clear: Contrastive learning for sentence representation.
\newblock \emph{ArXiv}, abs/2012.15466.

\bibitem[{Xie et~al.(2020{\natexlab{a}})Xie, Dai, Hovy, Luong, and
  Le}]{Xie2020UnsupervisedDA}
Qizhe Xie, Zihang Dai, Eduard Hovy, Thang Luong, and Quoc Le.
  2020{\natexlab{a}}.
\newblock Unsupervised data augmentation for consistency training.
\newblock In \emph{Proc. of NeurlPS}.

\bibitem[{Xie et~al.(2020{\natexlab{b}})Xie, Hovy, Luong, and
  Le}]{Xie2020SelfTrainingWN}
Qizhe Xie, E.~Hovy, Minh-Thang Luong, and Quoc~V. Le. 2020{\natexlab{b}}.
\newblock Self-training with noisy student improves imagenet classification.
\newblock In \emph{Proc. of CVPR}.

\bibitem[{Yang et~al.(2019)Yang, Dai, Yang, Carbonell, Salakhutdinov, and
  Le}]{yang2019xlnet}
Zhilin Yang, Zihang Dai, Yiming Yang, Jaime Carbonell, Russ~R Salakhutdinov,
  and Quoc~V Le. 2019.
\newblock Xlnet: Generalized autoregressive pretraining for language
  understanding.
\newblock In \emph{Proc. of NeurlPS}.

\bibitem[{Yarowsky(1995)}]{yarowsky-1995-unsupervised}
David Yarowsky. 1995.
\newblock \href {https://doi.org/10.3115/981658.981684} {Unsupervised word
  sense disambiguation rivaling supervised methods}.
\newblock In \emph{33rd Annual Meeting of the Association for Computational
  Linguistics}, pages 189--196, Cambridge, Massachusetts, USA. Association for
  Computational Linguistics.

\bibitem[{Yu et~al.(2018)Yu, Guo, Yi, Chang, Potdar, Cheng, Tesauro, Wang, and
  Zhou}]{yu-etal-2018-diverse}
Mo~Yu, Xiaoxiao Guo, Jinfeng Yi, Shiyu Chang, Saloni Potdar, Yu~Cheng, Gerald
  Tesauro, Haoyu Wang, and Bowen Zhou. 2018.
\newblock \href {https://doi.org/10.18653/v1/N18-1109} {Diverse few-shot text
  classification with multiple metrics}.
\newblock In \emph{Proceedings of the 2018 Conference of the North {A}merican
  Chapter of the Association for Computational Linguistics: Human Language
  Technologies, Volume 1 (Long Papers)}, pages 1206--1215, New Orleans,
  Louisiana. Association for Computational Linguistics.

\bibitem[{Zhang et~al.(2021)Zhang, He, Liu, Bing, and
  Li}]{Zhang2021BootstrappedUS}
Yan Zhang, Ruidan He, Zuozhu Liu, Lidong Bing, and Haizhou Li. 2021.
\newblock Bootstrapped unsupervised sentence representation learning.
\newblock In \emph{ACL/IJCNLP}.

\bibitem[{Zhu(2005)}]{Zhu2005SemiSupervisedLL}
Xiaojin Zhu. 2005.
\newblock Semi-supervised learning literature survey.
\newblock \emph{world}.

\bibitem[{Zoph et~al.(2020)Zoph, Ghiasi, Lin, Cui, Liu, Cubuk, and
  Le}]{Zoph2020RethinkingPA}
Barret Zoph, Golnaz Ghiasi, Tsung-Yi Lin, Yin Cui, Hanxiao Liu, Ekin~Dogus
  Cubuk, and Quoc Le. 2020.
\newblock Rethinking pre-training and self-training.
\newblock In \emph{Proc. of NeurlPS}.

\end{thebibliography}
\bibliographystyle{acl_natbib}

\end{document}